%% file: main.tex
\documentclass[sigconf, 9pt]{acmart}

\AtBeginDocument{%
  }

\setcopyright{none}
\renewcommand{\tableofcontents}{}
\renewcommand\footnotetextcopyrightpermission[1]{}
\usepackage{booktabs}       
\usepackage{multirow}       
\usepackage{graphicx}       
\usepackage{subcaption}     
\usepackage{xspace}         
\usepackage{xcolor}         
\usepackage{fontawesome5}   
\usepackage{listings}       
\usepackage{balance}        
\usepackage{afterpage}      

\hypersetup{
  colorlinks=true,
  linkcolor=ACMBlue,
  citecolor=ACMBlue,
  urlcolor=ACMBlue,
  filecolor=ACMBlue
}

\newcommand{\bench}{\textsc{HWE-Bench}\xspace}
\newcommand{\numcases}{417\xspace}
\newcommand{\numrepos}{six\xspace}
\newcommand{\hficon}{\raisebox{-0.22\height}{\includegraphics[height=3.1ex]{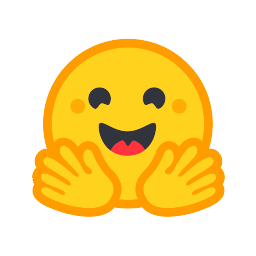}}}

\newcommand{\resourcebar}{%
  \LARGE
  \textbf{Code:}~{\faGithub}~\href{https://github.com/pku-liang/hwe-bench}%
    {\textcolor{ACMDarkBlue}{pku-liang/hwe-bench}}%
  \hspace{3em}%
  \textbf{Data:}~\hficon~\href{https://huggingface.co/datasets/henryen/hwe-bench}%
    {\textcolor{ACMDarkBlue}{henryen/hwe-bench}}%
  \par\vspace{1em}
}
\lstdefinestyle{promptblock}{
  basicstyle=\ttfamily\small,
  breaklines=true,
  breakatwhitespace=false,
  columns=fullflexible,
  keepspaces=true,
  showstringspaces=false,
  frame=single,
  xleftmargin=1em, xrightmargin=1em
}

\definecolor{diffadd}{RGB}{0,128,0}
\definecolor{diffdel}{RGB}{180,0,0}
\definecolor{diffhdr}{gray}{0.45}

\lstdefinelanguage{diff}{
  morecomment=[f][\color{diffadd}]{+},
  morecomment=[f][\color{diffdel}]{-},
  morecomment=[f][\color{diffhdr}]{@},
}

\begin{document}

\title{HWE-Bench: Benchmarking LLM Agents on Real-World Hardware Bug Repair Tasks}

\author{Fan Cui}
\affiliation{%
  \institution{Peking University}
  \city{Beijing}
  \country{China}}
\email{pku_cf@stu.pku.edu.cn}

\author{Hongyuan Hou}
\affiliation{%
  \institution{Peking University}
  \city{Beijing}
  \country{China}}
\email{houhy@stu.pku.edu.cn}

\author{Zizhang Luo}
\affiliation{%
  \institution{Peking University}
  \city{Beijing}
  \country{China}}
\email{semiwaker@pku.edu.cn}

\author{Chenyun Yin}
\affiliation{%
  \institution{Peking University}
  \city{Beijing}
  \country{China}}
\email{higgs@stu.pku.edu.cn}

\author{Yun Liang}
\authornote{Corresponding author.}
\affiliation{%
  \institution{Peking University}
  \city{Beijing}
  \country{China}}
\email{ericlyun@pku.edu.cn}

\renewcommand{\shortauthors}{Cui et al.}

\begin{teaserfigure}
  \centering
  \resourcebar
  \Description{Resource links to the HWE-Bench GitHub repository and Hugging Face dataset page.}
\end{teaserfigure}

\begin{abstract}
Existing benchmarks for hardware design primarily evaluate Large Language Models (LLMs) on isolated, component-level tasks such as generating HDL modules from specifications, leaving repository-scale evaluation unaddressed. We introduce \bench, the first large-scale, repository-level benchmark for evaluating LLM agents on real-world hardware bug repair tasks. \bench comprises \numcases task instances derived from real historical bug-fix pull requests across \numrepos major open-source projects spanning both Verilog/SystemVerilog and Chisel, covering RISC-V cores, SoCs, and security roots-of-trust. Each task is grounded in a fully containerized environment where the agent must resolve a real bug report, with correctness validated through the project's native simulation and regression flows. The benchmark is built through a largely automated pipeline that enables efficient expansion to new repositories. We evaluate seven LLMs with four agent frameworks and find that the best agent resolves 70.7\% of tasks overall, with performance exceeding 90\% on smaller cores but dropping below 65\% on complex SoC-level projects. We observe larger performance gaps across models than commonly reported on software benchmarks, and difficulty is driven by project scope and bug-type distribution rather than code size alone. Our failure analysis traces agent failures to three stages of the debugging process: fault localization, hardware-semantic reasoning, and cross-artifact coordination across RTL, configuration, and verification components, providing concrete directions for developing more capable hardware-aware agents.
\end{abstract}

\maketitle


\input{src/01_intro}
\input{src/02_related}
\input{src/03_benchmark}
\input{src/04_evaluation}
\input{src/05_conclusion}

\balance
\bibliographystyle{ACM-Reference-Format}
\bibliography{references}

\end{document}

%% file: src/01_intro.tex

\section{Introduction}

The potential of Large Language Models (LLMs) in hardware design and Electronic Design Automation (EDA) has attracted growing interest~\citep{edasurvey}. Existing benchmarks such as VerilogEval~\citep{verilogeval} and RTLLM~\citep{rtllm} have driven progress in hardware-oriented models including RTLCoder~\citep{rtlcoder}, CodeV~\citep{codev}, and OriGen~\citep{origen}, but focus predominantly on component-level code generation from natural-language specifications. CVDP~\citep{cvdp} advances beyond generation to encompass verification, debugging, and agentic interaction, yet its tasks are individual design problems isolated from their project context, without requiring the agent to navigate a full repository or invoke native verification flows. None of these benchmarks addresses the complexities of repository-level engineering, where real-world hardware projects demand reasoning across heterogeneous artifacts including RTL source, verification components, IP configurations, and build scripts, as well as interaction with project-native build and simulation flows. A fundamental question therefore remains open: can LLM-based agents perform repository-level bug repair within the full context of a real hardware project?

Repository-level, execution-grounded evaluation has driven substantial progress in software engineering, where benchmarks such as SWE-bench~\citep{swebench} and its extensions~\citep{swebenchpro, terminalbench} enabled standardized, reproducible evaluation of agent capabilities. No comparable infrastructure exists for hardware, and building it poses distinct challenges. Open-source HDL repositories are scarce: Verilog and SystemVerilog comprise 100--1,000$\times$ fewer files than Python in standard code corpora~\citep{stackv2}, let alone hardware DSLs such as Chisel. Even when suitable repositories are identified, constructing reproducible, verifiable benchmark instances remains difficult: where software benchmarks can rely on standardized build and test infrastructure, hardware verification depends on heterogeneous, often proprietary toolchains, and many bugs lack automated regression tests. Establishing reproducible validation for each task instance therefore demands substantial manual effort.

\begin{figure*}[t]
    \centering
    \includegraphics[width=0.8\textwidth]{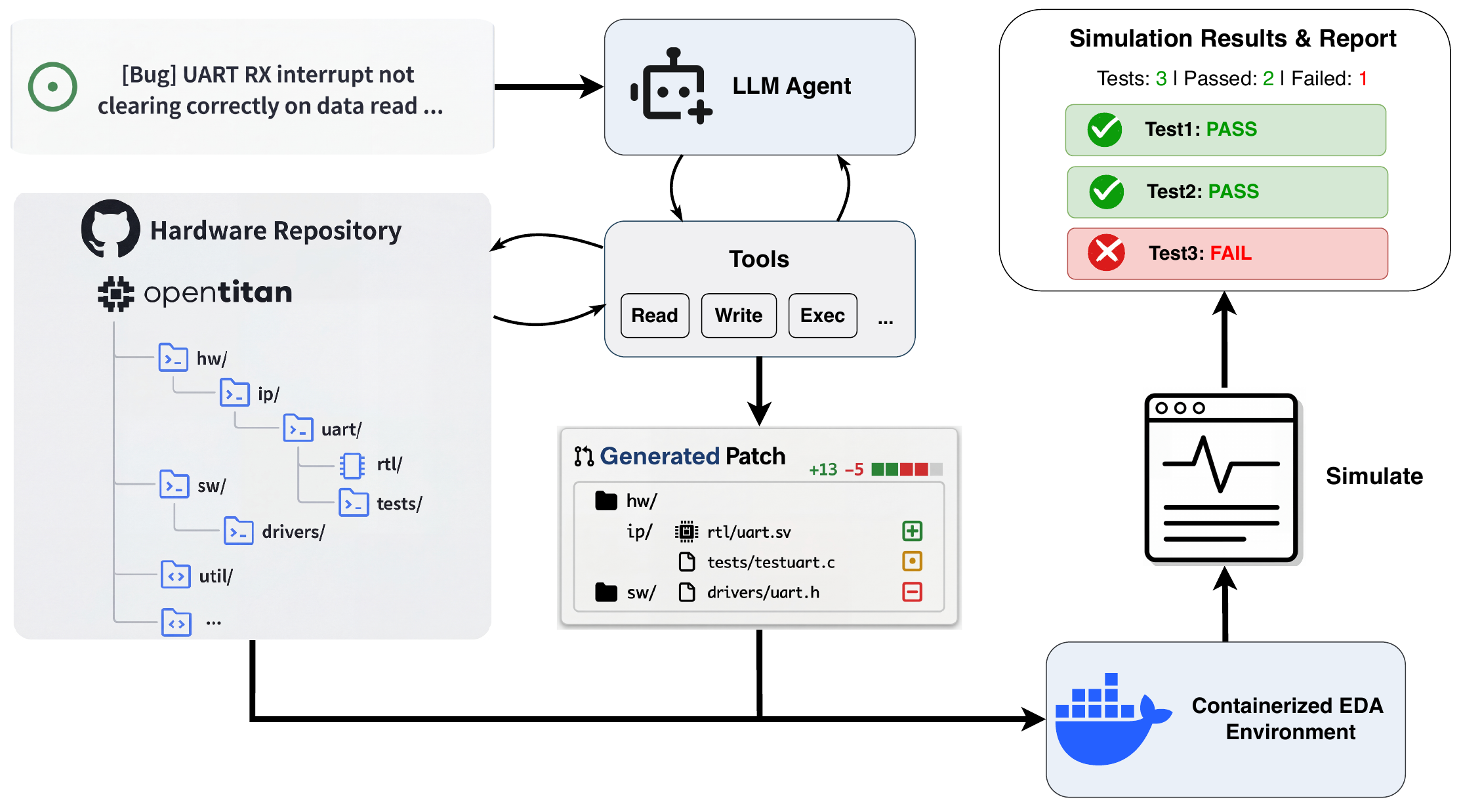}
    \caption{Overview of a task instance in \bench.}
    \Description{Overview diagram of the HWE-Bench task workflow.}
    \label{fig:overview}
\end{figure*}

To address this gap, we introduce \bench, the first large-scale, repository-level benchmark for evaluating LLM agents on real-world hardware bug repair tasks. As illustrated in \autoref{fig:overview}, each task instance provides an LLM agent with a bug report and the full hardware repository; the agent interacts with the codebase through tools such as file reading, editing, and shell execution to produce a patch, which is then validated through the project's native simulation suite in a containerized environment. \bench comprises \numcases task instances curated from \numrepos major open-source projects (OpenTitan, Caliptra, XiangShan, Ibex, CVA6, and Rocket Chip), spanning both traditional HDLs (Verilog/SystemVerilog) and a hardware DSL (Chisel), with diverse design categories from RISC-V cores to system-on-chip platforms and security roots-of-trust. Each task is grounded in a fully containerized, execution-based evaluation environment. The benchmark is constructed through a systematic, largely automated pipeline that filters genuine hardware bug fixes from repository noise and establishes reproducible, containerized validation for each instance.

We evaluate seven LLMs, both proprietary and open-source, using four agent frameworks across all six repositories. The best-performing agent achieves an overall resolved rate of 70.7\%, but performance varies considerably: notably, we observe larger performance gaps between proprietary and open-source models than commonly reported on software benchmarks, suggesting that hardware tasks expose capability differences that software-focused evaluations do not capture. Across repositories, we observe a clear difficulty gradient governed primarily by project scope and bug-type distribution rather than code size alone, with resolved rates ranging from above 90\% on smaller projects to below 65\% on the largest SoC-level project. Our failure analysis identifies three recurring failure modes, fault localization, hardware-semantic reasoning, and cross-artifact coordination, as well as distinctive model-specific behavioral differences and repository-specific challenges.

In summary, our contributions are as follows:
\begin{itemize}
  \item We present \bench, the first repository-level, execution-grounded benchmark for hardware bug repair, comprising \numcases tasks across \numrepos repositories spanning both traditional HDLs (Verilog/SystemVerilog) and a hardware DSL (Chisel).
  \item We develop a largely automated benchmark construction pipeline, enabling efficient expansion to new hardware repositories.
  \item We evaluate seven LLMs with four agent frameworks across all repositories, revealing larger capability gaps between proprietary and open-source models than observed on software benchmarks, and identifying hardware-specific failure modes that inform future agent design.
\end{itemize}

%% file: src/02_related.tex

\section{Related Work}

\begin{table*}[t]
\centering
\caption{Comparison of benchmarks for hardware and software engineering tasks.}
\label{tab:benchmark_comparison}
\resizebox{0.9\textwidth}{!}{%
\begin{tabular}{@{}lllllll@{}}
\toprule
\textbf{Benchmark} & \textbf{Domain} & \textbf{Level} & \textbf{Task Type} & \textbf{Languages} & \textbf{Verification} & \textbf{Environment} \\
\midrule
VerilogEval~\citep{verilogeval}   & Hardware & Component  & Code Generation   & Verilog/SV        & Simulation        & Isolated \\
RTLLM~\citep{rtllm}               & Hardware & Component  & Code Generation   & Verilog/SV        & Simulation        & Isolated \\
HWFixBench~\citep{hwfixbench}     & Hardware & File       & Bug Repair        & Verilog/SV        & LLM-as-Judge      & Isolated \\
CVDP~\citep{cvdp}                 & Hardware & Component  & Multi-task        & Verilog/SV        & Simulation        & Isolated \\
\midrule
SWE-bench~\citep{swebench}        & Software & Repository & Bug Repair        & Python            & Unit Tests        & Containerized \\
SWE-bench Pro~\citep{swebenchpro} & Software & Repository & Bug Repair        & Multi-language    & Unit Tests        & Containerized \\
\midrule
\textbf{\bench (Ours)}            & \textbf{Hardware} & \textbf{Repository} & \textbf{Bug Repair} & \textbf{Verilog/SV + Chisel} & \textbf{Simulation} & \textbf{Containerized} \\
\bottomrule
\end{tabular}%
}
\end{table*}

\paragraph{Hardware Engineering Benchmarks.}
Evaluation of LLMs for hardware has progressed from code generation to broader engineering tasks, as shown in \autoref{tab:benchmark_comparison}. Early benchmarks such as VerilogEval~\citep{verilogeval} and RTLLM~\citep{rtllm} focused on generating individual modules from specifications, providing simulation-based correctness checks at the component level. CVDP~\citep{cvdp} broadened both the task scope (adding verification, debugging, and agentic formats) and scale (783 problems), while HWFixBench~\citep{hwfixbench} moved toward bug repair using real repository data. However, CVDP tasks are problem-scoped rather than repository-scoped: even in agentic settings, agents do not operate over the original project repository or its native regression flow. HWFixBench provides file-level inputs with LLM-based semantic matching rather than execution-based validation. \bench addresses both limitations by embedding tasks in full repositories and validating patches through native simulation in containerized environments.

\paragraph{Software Engineering Benchmarks.}
Repository-level, execution-grounded evaluation was established most visibly in software engineering with SWE-bench~\citep{swebench}, which demonstrated that grounding agent evaluation in real GitHub issues with unit-test validation provides a reproducible and scalable evaluation methodology. This paradigm has since been extended to multiple languages and longer-horizon tasks~\citep{multiswebench, swebenchpro}, as well as terminal-based workflows~\citep{terminalbench}. \bench brings this evaluation paradigm to hardware, but requires fundamentally different infrastructure to handle heterogeneous EDA toolchains and non-standardized verification flows.

\paragraph{Automated RTL Repair.}
Traditional automated repair tools for hardware, such as CirFix~\citep{cirfix} and RTL-Repair~\citep{rtlrepair}, also target hardware bug fixing but under a much narrower formulation: they operate on a specific module with a given failing testbench and search for minimal patches that satisfy the provided tests. The task formulation in \bench is substantially broader: agents receive only a natural-language bug report and the full repository, and must independently localize the fault, navigate the codebase, and invoke project-native build and simulation flows to validate their fix.

%% file: src/03_benchmark.tex

\section{The \bench Benchmark}
\label{sec:benchmark}

\subsection{Construction Pipeline}

\begin{figure*}[t]
    \centering
    \includegraphics[width=0.95\textwidth]{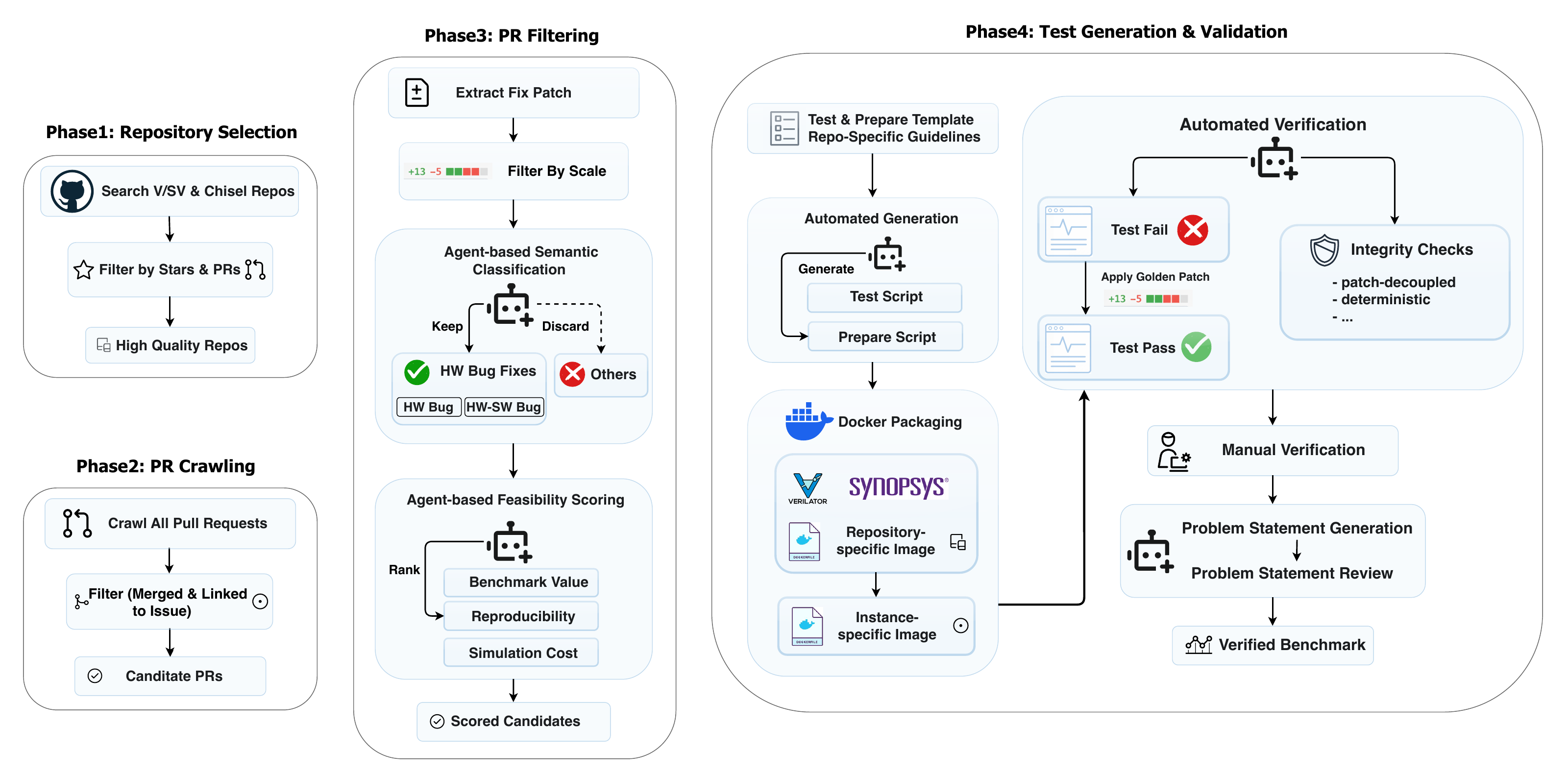}
    \caption{Construction pipeline of \bench.}
    \Description{Pipeline diagram showing four phases: repository selection, PR crawling, semantic filtering, and validation with environment construction.}
    \label{fig:pipeline}
\end{figure*}

The construction of \bench follows a multi-phase pipeline (\autoref{fig:pipeline}) designed to extract verifiable bug repair tasks from real-world hardware projects.

\paragraph{Repository Selection.}
The open-source hardware landscape offers far fewer well-maintained repositories than the software ecosystem. We search GitHub for projects using Verilog, SystemVerilog, or Chisel with a substantial development history (over 400 pull requests) and over 100 stars. From the small pool of candidates meeting these criteria, we select six repositories that collectively represent diverse hardware design categories: \textbf{OpenTitan}, an open-source Root-of-Trust SoC; \textbf{Caliptra}, a security-focused Root-of-Trust IP; \textbf{XiangShan}, a high-performance out-of-order RISC-V processor; \textbf{Ibex}, a production-quality 32-bit RISC-V core; \textbf{CVA6}, a 6-stage RISC-V core from the CORE-V family; and \textbf{Rocket Chip}, a parameterizable RISC-V SoC generator. These span both traditional HDLs (Verilog/SystemVerilog) and a hardware DSL (Chisel), covering a range of design complexities, verification methodologies, and application domains.

\paragraph{PR Collection and Filtering.}
We collect all pull requests from each repository and retain those that are merged into the main branch and linked to at least one GitHub issue, ensuring each candidate addresses a clearly defined problem accepted by the project's maintainers~\citep{swebench, multiswebench}. However, hardware repositories present a challenge not encountered in software: a large fraction of merged PRs address activities peripheral to hardware debugging, such as documentation updates, CI/CD configurations, build script changes, and EDA toolchain upgrades. We apply two stages of filtering to address these challenges. A scale filter excludes large-scale refactorings and feature additions, while preserving the multi-file complexity characteristic of hardware bug fixes. A semantic filter, implemented as an LLM-based classification agent, then analyzes each remaining PR to retain only genuine hardware-related bug fixes, including both RTL design bugs and hardware-software interaction bugs, while discarding the non-design activity described above. An automated feasibility scoring stage further evaluates each candidate along multiple dimensions, including representativeness as a debugging challenge, strength of available reproduction clues, and expected simulation cost, to prioritize PRs most suitable for benchmark inclusion.

\paragraph{Test Generation and Validation.}
A key advance over prior benchmark construction methodology, which can require large teams of human annotators~\citep{multiswebench}, is our largely automated test generation and validation pipeline. We define a common test script template that standardizes the interface (environment setup, simulation invocation, and PASS/FAIL output parsing), then for each repository provide repository-specific guidelines describing its toolchain, verification flow, and simulation conventions. An LLM agent generates a test script and an environment preparation script for each candidate PR, following the template when the project's native verification infrastructure allows it and producing custom implementations when it does not.

The resulting environments are packaged as Docker images: a repository-level base image containing EDA tools and project dependencies, and per-instance images that apply the base commit, environment patches, and test scripts. Each instance then undergoes automated verification: the pipeline runs the test on the base commit (expecting FAIL), applies the ground-truth patch, and re-runs the test (expecting PASS). Beyond this FAIL-to-PASS check, the verification stage also ensures that the test is decoupled from the specific implementation of the ground-truth patch, so that alternative correct fixes are not rejected. Instances that fail verification are iteratively repaired or excluded. All surviving instances then undergo manual review to catch issues that automated checks cannot reliably detect, such as tests that pass for the wrong reason or environment configurations that are subtly non-reproducible.

Finally, as prior work has shown that raw issue descriptions are often too unstructured or incomplete for reliable evaluation~\citep{swebenchpro}, we generate a self-contained problem statement for each instance using a separate LLM agent. Since the generated text may inadvertently reveal fix approaches, specific file paths, or simulation commands not present in the original issue, an automated review step checks for such information leakage, as well as semantic completeness and bidirectional alignment with the test scope. The agent thus receives this generated problem statement, rather than the raw issue text, as its task input.

\subsection{Benchmark Overview}

\begin{table}[t]
\centering
\caption{Summary of repositories in \bench. LoC counts all code in the repository (excluding submodules).}
\label{tab:repo_overview}
\resizebox{\columnwidth}{!}{%
\begin{tabular}{@{}llllrl@{}}
\toprule
\textbf{Repository} & \textbf{HDL} & \textbf{Design Type} & \textbf{Simulator} & \textbf{Tasks} & \textbf{LoC} \\
\midrule
OpenTitan    & V/SV   & SoC / RoT          & VCS       & 245 & 1,540K \\
Ibex         & V/SV   & RISC-V Core        & Verilator & 35  & 551K \\
CVA6         & V/SV   & RISC-V Core        & Verilator & 35  & 403K \\
Caliptra     & V/SV   & Security RoT       & Verilator & 16  & 364K \\
XiangShan    & Chisel & OoO Processor      & Verilator & 54  & 117K \\
Rocket Chip  & Chisel & RISC-V SoC Gen.    & Verilator & 32  & 47K \\
\midrule
\textbf{Total} & & & & \textbf{\numcases} & \\
\bottomrule
\end{tabular}%
}
\end{table}

Starting from over 30,000 pull requests across the six repositories, the pipeline yields \numcases verified instances after all filtering, validation, and manual review stages. \autoref{tab:repo_overview} summarizes the six repositories in \bench. The benchmark spans a range of design complexity, from small, well-structured cores (Ibex, CVA6) to large-scale projects with sophisticated verification infrastructure (OpenTitan, XiangShan). OpenTitan contributes the largest share of tasks, reflecting both its extensive pull request history and the richness of its DV/UVM verification flows; the remaining five repositories provide diversity in HDL, design style, and complexity. Five repositories use the open-source Verilator simulator, while OpenTitan relies on Synopsys VCS to support its UVM-based verification environment. Users without commercial licenses can evaluate on the Verilator-based subset (172 of \numcases tasks).

\subsection{Task Properties}

\begin{figure*}[ht]
\centering
\begin{subfigure}[b]{0.48\textwidth}
    \includegraphics[width=\textwidth]{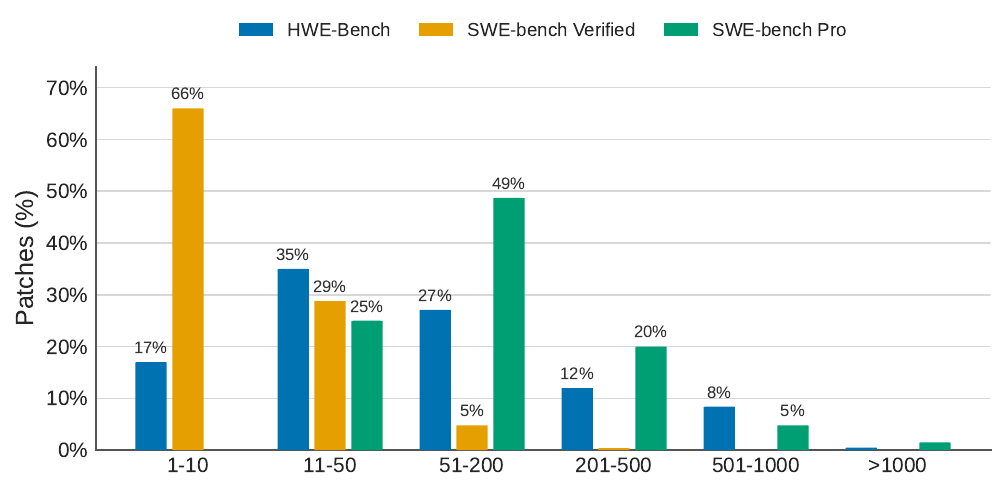}
    \caption{Lines changed per patch.}
    \label{fig:patch_lines}
\end{subfigure}
\hfill
\begin{subfigure}[b]{0.48\textwidth}
    \includegraphics[width=\textwidth]{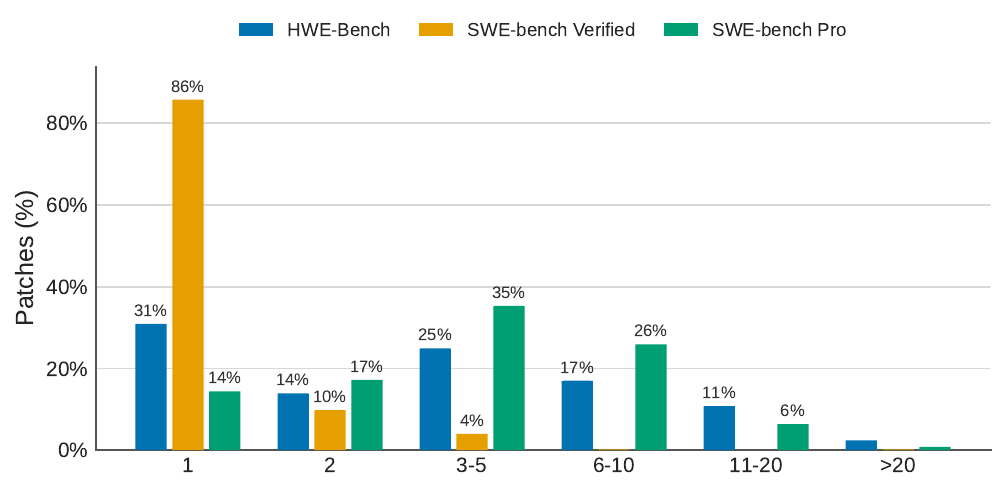}
    \caption{Files modified per patch.}
    \label{fig:patch_files}
\end{subfigure}
\caption{Distribution of ground-truth patches in \bench compared with SWE-bench Verified~\citep{swebench} and SWE-bench Pro~\citep{swebenchpro}.}
\Description{Two histograms comparing patch size distributions across HWE-Bench, SWE-bench Verified, and SWE-bench Pro.}
\label{fig:patch_distribution}
\end{figure*}

\begin{figure}[t]
    \centering
    \includegraphics[width=\columnwidth]{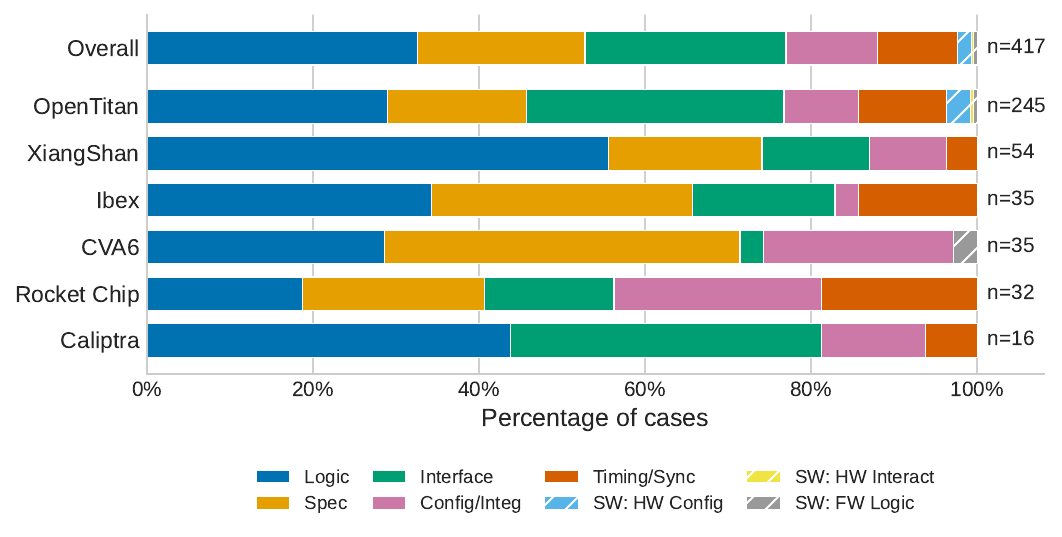}
    \caption{Distribution of bug categories across the six repositories in \bench, classified by the semantic filter.}
    \Description{Horizontal stacked bar chart showing bug category breakdown for HWE-Bench overall and per repository.}
    \label{fig:bug_categories}
\end{figure}

\paragraph{Patch Complexity.}
\autoref{fig:patch_distribution} situates \bench relative to SWE-bench Verified~\citep{swebench} and SWE-bench Pro~\citep{swebenchpro}. The contrast is informative: Verified is dominated by small, single-file fixes, whereas Pro places much more mass on large patches that span many files. \bench shows a distinct profile. Its patches are frequently multi-file and materially larger than those in Verified, indicating that tasks typically require coordinated edits rather than localized line changes. At the same time, the distribution is not driven by the largest patch regime seen in Pro; instead, it is centered on medium-scale changes that demand cross-file reasoning without collapsing into a few extreme cases.

This shape is well aligned with hardware development practice. In RTL repositories, a bug fix commonly propagates across design logic, parameterization or configuration, and verification components such as testbenches, scoreboards, or UVM infrastructure. \bench reflects this engineering structure directly, yielding a benchmark that is both realistic and discriminative for evaluating agents on coordinated, repository-scale hardware fixes.

\paragraph{Bug Categories.}
\autoref{fig:bug_categories} shows the distribution of bug categories across the six repositories in \bench. The taxonomy organizes hardware bugs by their dominant failure mode into functional logic errors, specification deviations, interface and protocol violations, timing and synchronization issues, and configuration or integration errors, with a smaller secondary group for software and firmware bugs that misuse the hardware. The benchmark is dominated by hardware design bugs, with all five root-cause classes well represented; logic errors form the largest category, and interface/protocol and timing bugs, phenomena characteristic of concurrent hardware execution, together contribute a substantial share. The small number of software/firmware cases appears almost entirely in OpenTitan, reflecting its extensive firmware and driver layers.

The distribution also varies meaningfully across repositories. Specification-related bugs dominate in CVA6 and Ibex, where strict RISC-V ISA adherence is the primary correctness criterion, while XiangShan is dominated by logic errors arising from its complex out-of-order pipeline. OpenTitan shows the most diverse profile, with interface and configuration bugs prevalent due to its multi-IP SoC architecture. This diversity ensures that \bench evaluates agents across a broad range of reasoning challenges encountered in real hardware development.

%% file: src/04_evaluation.tex
\section{Evaluation}

\subsection{Experimental Setup}

\paragraph{Models.}
We evaluate seven state-of-the-art LLMs spanning both proprietary and open-source families. The proprietary models include GPT-5.4~\citep{gpt54} (configured with the \texttt{xhigh} reasoning effort), Claude Opus 4.6~\citep{claudeopus46} and Claude Sonnet 4.6~\citep{claudesonnet46} (both configured with the \texttt{high} thinking effort). The open-source models include GLM 5.1~\citep{glm51}, Qwen3.6 Plus~\citep{qwen36plus}\footnote{Qwen3.6 Plus is the hosted flagship of the Qwen3.6 family, whose open-weight variants have been announced but not yet released; we include it as an upper-bound reference for the forthcoming open-source models.}, DeepSeek V3.2~\citep{deepseekv32}, and Kimi K2.5~\citep{kimik25}.

\paragraph{Agent Frameworks.}
Each model requires an agent scaffolding to interact with the containerized environment and invoke EDA tools. To reflect how these models are actually used in practice, we pair each model with its official CLI framework when available: Claude Opus/Sonnet 4.6 with Claude Code~\citep{claudecode}, GPT-5.4 with Codex CLI~\citep{codexcli}, and Kimi K2.5 with Kimi CLI~\citep{kimicli}, all accessed through their bundled coding-plan subscriptions. The remaining models (DeepSeek V3.2, Qwen3.6 Plus, GLM 5.1) are evaluated with OpenHands~\citep{openhands} through direct API calls.\footnote{Qwen3.6 Plus has an official CLI Qwen Code but performed noticeably worse than with OpenHands in our pilot comparison; we therefore use OpenHands for this model.} This setup reflects the dominant deployment path for each model; results should therefore be interpreted as end-to-end agent-system performance rather than isolated model capabilities.

\paragraph{Evaluation Protocol.}
Each agent solves every task independently in its containerized environment, under a uniform execution budget of a 6-hour wall-clock timeout and 500 iterations. The agent receives the self-contained problem statement together with the full repository context, with no hints about the ground-truth patch. After the agent terminates, an evaluation script runs the verification test on its final patch and produces a binary PASS/FAIL signal.

\paragraph{Metrics.}
We report two complementary metrics. \textbf{Resolved Rate} measures end-to-end success: the fraction of tasks for which the agent's generated patch causes the verification test to transition from FAIL to PASS. \textbf{File-Level Precision} measures patch focus: the fraction of files modified by the agent that also appear in the ground-truth patch. A high precision primarily reflects accurate fault localization, indicating that the agent identified and modified the correct files, and secondarily that it avoided modifying unrelated files during the solving process.

\subsection{Main Results}

\begin{figure*}[t]
\centering
\begin{minipage}[b]{0.62\textwidth}
    \centering
    \includegraphics[width=\textwidth]{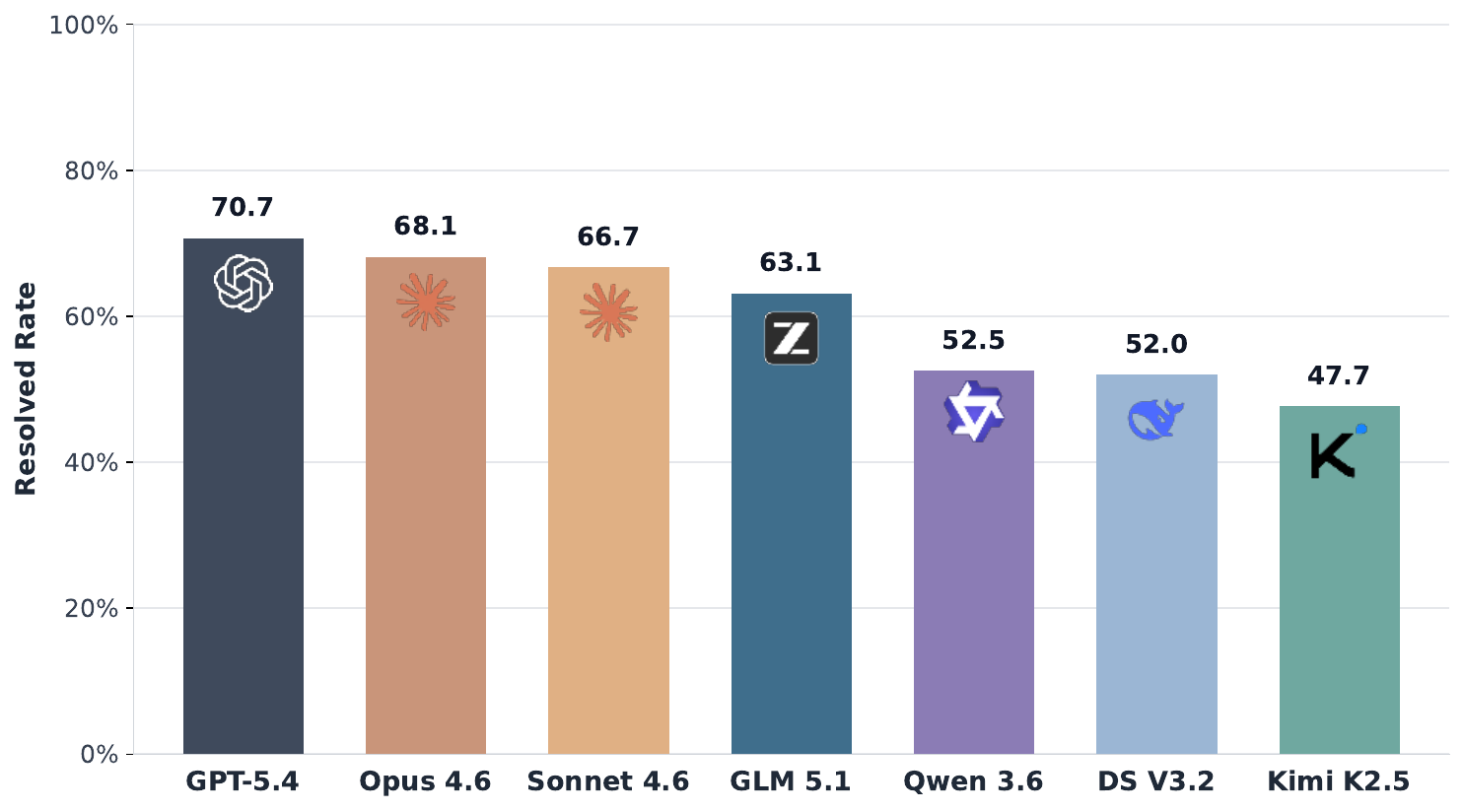}
    \subcaption{\bench}
    \label{fig:resolved_hwe}
\end{minipage}
\hfill
\begin{minipage}[b]{0.36\textwidth}
    \centering
    \includegraphics[width=\textwidth]{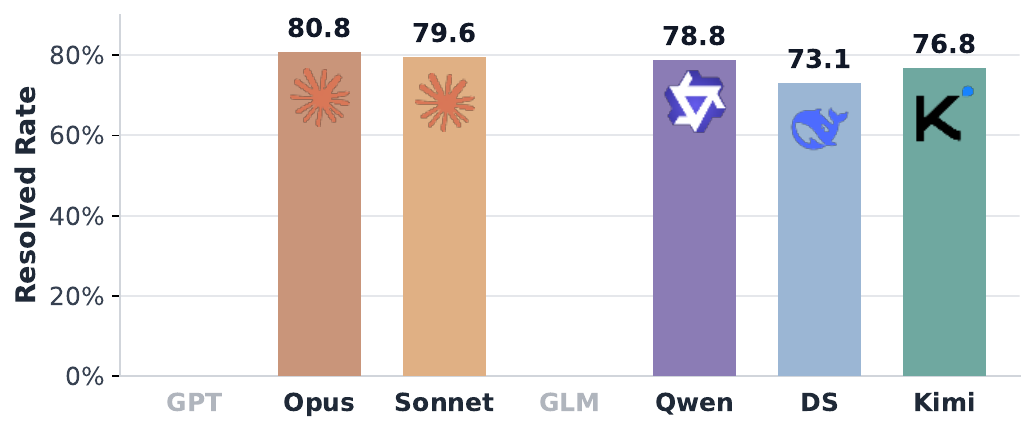}
    \subcaption{SWE-bench Verified}
    \label{fig:resolved_verified}
    \vspace{0.5em}
    \includegraphics[width=\textwidth]{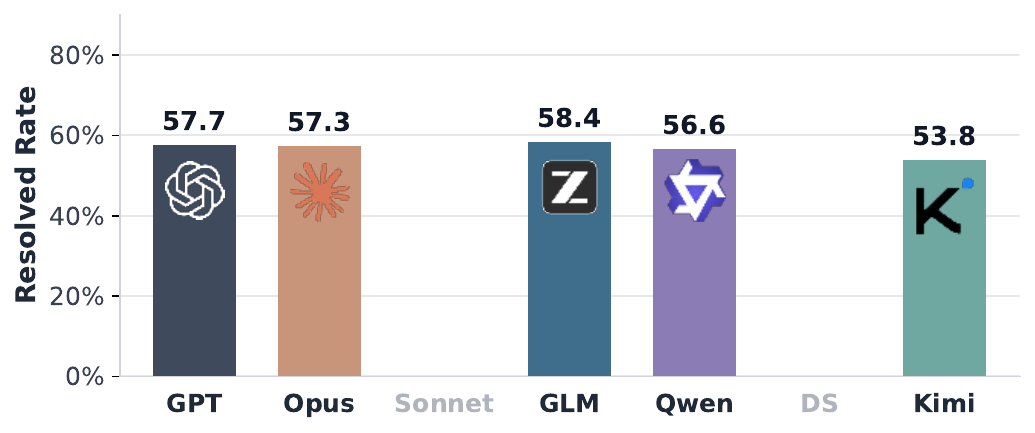}
    \subcaption{SWE-bench Pro}
    \label{fig:resolved_pro}
\end{minipage}
\caption{Overall resolved rate of the same set of models across three benchmarks. Scores in (b) are from each model's official technical report or release blog~\citep{claudeopus46, claudesonnet46, qwen36plus, deepseekv32, kimik25}; scores in (c) are from GLM-5.1's model card~\citep{glm51}. Faded bars indicate models without officially reported scores.}
\Description{Three bar charts comparing resolved rates across HWE-Bench, SWE-bench Verified, and SWE-bench Pro.}
\label{fig:overall_resolved}
\end{figure*}

\begin{table*}[t]
\centering
\caption{Resolved Rate per repository with patch file-level precision. Cells report count (\%).}
\label{tab:main_results}
\resizebox{\textwidth}{!}{%
\begin{tabular}{@{}l cccccc c c@{}}
\toprule
 & \multicolumn{7}{c}{\textbf{Resolved Rate}} & \\
\cmidrule(lr){2-8}
\textbf{Model} & \textbf{OpenTitan (245)} & \textbf{Ibex (35)} & \textbf{CVA6 (35)} & \textbf{Caliptra (16)} & \textbf{XiangShan (54)} & \textbf{Rocket Chip (32)} & \textbf{Overall (417)} & \textbf{Precision} \\
\midrule
\multicolumn{9}{l}{\textit{Proprietary}} \\
GPT-5.4 xhigh (Codex CLI)              & \textbf{157 (64.1)} & \textbf{32 (91.4)} & \textbf{34 (97.1)} & \textbf{15 (93.8)} & \textbf{37 (68.5)} & \textbf{20 (62.5)} & \textbf{295 (70.7)} & 0.619 \\
Claude Opus 4.6 high (Claude Code)      & 156 (63.7) & 29 (82.9) & 30 (85.7) & 14 (87.5) & 36 (66.7) & 19 (59.4) & 284 (68.1) & \textbf{0.928} \\
Claude Sonnet 4.6 high (Claude Code)    & 152 (62.0) & 29 (82.9) & 30 (85.7) & 13 (81.3) & 36 (66.7) & 18 (56.3) & 278 (66.7) & 0.916 \\
\midrule
\multicolumn{9}{l}{\textit{Open-Source}} \\
GLM 5.1 (OpenHands)                     & 152 (62.0) & 26 (74.3) & 28 (80.0) & 11 (68.8) & 29 (53.7) & 17 (53.1) & 263 (63.1) & 0.892 \\
Qwen3.6 Plus (OpenHands)                & 118 (48.2) & 22 (62.9) & 30 (85.7) & 9 (56.3) & 25 (46.3) & 15 (46.9) & 219 (52.5) & 0.878 \\
DeepSeek V3.2 (OpenHands)               & 116 (47.3) & 22 (62.9) & 24 (68.6) & 13 (81.3) & 28 (51.9) & 14 (43.8) & 217 (52.0) & 0.865 \\
Kimi K2.5 (Kimi CLI)                    & 106 (43.3) & 19 (54.3) & 28 (80.0) & 10 (62.5) & 22 (40.7) & 14 (43.8) & 199 (47.7) & 0.859 \\
\bottomrule
\end{tabular}%
}
\end{table*}

\autoref{tab:main_results} and \autoref{fig:overall_resolved} present the main evaluation results. GPT-5.4 xhigh achieves the highest overall resolved rate at 70.7\%, including 22 cases that no other model resolves. Claude Opus 4.6 (68.1\%) and Sonnet 4.6 (66.7\%) follow closely. Among open-source models, GLM 5.1 stands out at 63.1\%, closely trailing the proprietary models and substantially outperforming the remaining open-source entries. The gap between the top proprietary and top open-source model is modest (7.6\%), but widens considerably toward the lower end of the ranking, with Kimi K2.5 at 47.7\%. At the per-repository level, proprietary models already resolve over 80\% of tasks on smaller repositories (CVA6, Ibex, Caliptra), indicating strong performance on well-structured, moderately sized hardware projects. At the same time, 45 of the \numcases cases (10.8\%) remain unsolved by all seven models. These cases typically involve deep protocol semantics combined with cross-module coordination, pointing to causal reasoning across abstraction layers as the key missing capability.

\autoref{fig:overall_resolved} also places \bench alongside SWE-bench Verified and SWE-bench Pro. The comparison reveals two observations. First, model performance diverges more sharply on hardware tasks: on SWE-bench Verified, all models cluster between 73\% and 81\%, with open-source models such as Qwen3.6 Plus and Kimi K2.5 approaching proprietary ones. On \bench, the same models fall to 52.5\% and 47.7\% while proprietary models remain above 66\%, widening the performance spread from under 8\% to over 23\%. A similar compression appears on SWE-bench Pro, where models again cluster within a narrow band. GLM 5.1 is consistent with this trend: although it drops from leading on SWE-bench Pro to fourth on \bench, it still heads the open-source group at 63.1\% and remains well separated from the remaining three open-source models (47--52\%). Hardware tasks thus expose performance differences that software-focused evaluations compress. Second, no model achieves near-saturated performance on \bench, indicating that it poses a meaningful and lasting challenge for current agents.

The File-Level Precision column in \autoref{tab:main_results} shows that, for most models, precision correlates positively with resolved rate. Claude Opus 4.6 achieves both the highest precision (0.928) and the second-highest resolved rate; GLM 5.1 likewise pairs its strong resolved rate with good precision (0.892). Lower precision can stem from multiple causes: inaccurate fault localization leading to edits in unrelated modules, over-broad modifications to files adjacent to the fix, or poor cleanup of intermediate artifacts (scratch scripts, backup files) generated during iterative debugging.

GPT-5.4 xhigh is a notable exception to this pattern: it achieves the highest resolved rate but the lowest precision (0.619). To understand this, we examined its generated patches and trajectories. The extra files are not workspace noise but real repository collateral: template and auto-generated mirrors, DV test infrastructure, and configuration files that are semantically related to the fix but absent from the ground-truth patch. This suggests that GPT-5.4 applies a broader, system-level understanding of the repository, proactively updating related components beyond the patch, at the cost of a less focused diff.

\subsection{Cross-Repository Difficulty Analysis}
\label{sec:cross_repo}

The per-repository breakdown in \autoref{tab:main_results} reveals a clear difficulty gradient. CVA6, Ibex, and Caliptra form an easier tier where even the weakest model resolves over half the cases and the best model exceeds 90\%. XiangShan and OpenTitan are substantially harder, with resolved rates dropping by 20--30 percentage points across all models. Rocket Chip, despite having the smallest codebase among the six repositories (47K LoC), falls into the hard tier alongside projects an order of magnitude larger.

This ordering suggests that repository size alone does not fully explain the difficulty gradient. OpenTitan (1.5M LoC) and Rocket Chip (47K LoC) differ by more than 30$\times$ in code volume, yet the best model achieves similar resolved rates on both (64.1\% vs.\ 62.5\%). Two factors better explain the observed gradient. First, project scope: OpenTitan and XiangShan are full-scale SoC and processor designs whose bug fixes frequently involve multi-module interactions and sophisticated verification flows (DV/UVM, difftest), whereas the easier repositories are smaller, more self-contained cores with simpler verification setups. Second, and more subtle, is the distribution of bug types. Rocket Chip has an unusually high proportion of configuration and integration bugs (over 40\% of its cases, compared to 20--25\% in other repositories). These bugs are particularly challenging because the symptom often points to a different module than the one that needs to be fixed, causing systematic fault-localization failures.

Concrete examples illustrate this pattern. In one Rocket Chip case, a clock-domain issue in the BootROM leads most agents to patch only that module, but the ground-truth fix requires restructuring the control-peripheral path across five interconnected components. In another, an apparent APB fanout decode error draws agents to patch the fanout logic, while the actual fix adds an error-handling path in the debug subsystem and tightens the APB address range. In both cases, agents that rely on keyword or proximity-based localization consistently edit the wrong site. By contrast, XiangShan's solved cases are predominantly local logic fixes within a single module, where the symptom and root cause coincide, making localization straightforward despite the repository's large overall size.

Since \bench spans both Verilog/SystemVerilog and Chisel, a natural question is whether the HDL itself affects difficulty. The two most complex projects in the benchmark, OpenTitan (V/SV) and XiangShan (Chisel), show comparable resolved rates across all models (e.g., 64.1\% vs.\ 68.5\% for the best agent), suggesting that the choice of HDL has limited impact relative to the project-level factors discussed above.

\subsection{Failure Analysis}

To understand why agents fail on \bench, we examined trajectories and generated patches across all seven models on a broad sample of failed cases. We first identify three recurring failure modes that account for the majority of unsuccessful attempts, then discuss model-specific behavioral differences and repository-specific challenges.

\paragraph{Localization Failure.}
The first stage at which agents fail is fault localization: identifying which files and modules need to be modified. In a significant fraction of failed cases, the agent edits the file where the failure manifests rather than the file that owns the bug. This occurs most frequently in exception and debug handling paths, address decoding logic, and cross-module integration wrappers. On Ibex, agents targeting an illegal-instruction reporting bug consistently edit the controller (where the symptom appears in simulation logs) rather than the fetch stage (where the instruction data is incorrectly forwarded). On XiangShan, a TLB address-width bug leads agents to rewrite the TLB module itself, while the ground-truth fix is a narrow change in a package-level definition upstream. Agents that rely on keyword matching between the issue description and the codebase are particularly susceptible, as the symptom description naturally references the module where the error is observed rather than where it originates.

\paragraph{Reasoning Failure.}
Even when agents correctly localize the bug, they frequently produce patches that are syntactically valid but fail to respect the underlying hardware contract. Common manifestations include incomplete state-machine transitions that leave stale state after pipeline flush or replay, partial handshake propagation where one side of a valid-ready interface is updated but the complementary side is not, and CSR or trap-handling fixes that address the register surface without fully repairing the control-flow redirection logic. On XiangShan, for example, agents correctly identify a dependency-checking module as the fix site but tighten the wrong index predicate (e.g., modifying \texttt{sqIdx} checks when the ground-truth fix redefines the dependency around \texttt{robIdx}), producing a patch that compiles but violates the pipeline's ordering contract. These failures suggest that current models can navigate hardware codebases effectively but lack the deep understanding of concurrent, cycle-level behavior needed to produce correct fixes.

\paragraph{Coordination Failure.}
Hardware bug fixes frequently require coordinated edits across multiple layers: RTL source, configuration files (e.g., HJSON register descriptions in OpenTitan), auto-generated collateral, and verification components. Even when agents produce a correct RTL fix, they consistently fail to propagate the change to these coupled artifacts, causing the verification test to remain in a failing state. For example, on an OpenTitan peripheral register interface bug, multiple agents correctly fix the RTL logic but omit the corresponding HJSON register description and auto-generated mirror updates. This pattern is especially common in repositories with tightly coupled design and verification artifacts, such as OpenTitan and Caliptra, and represents a challenge less common in standard software bug-repair benchmarks, where fixes rarely require synchronized edits to generated code or declarative configuration.

\paragraph{Model-Specific Observations.}
Beyond these shared patterns, individual models exhibit distinctive failure characteristics shaped by their debugging strategies. The Claude models (Opus and Sonnet) operate almost entirely through static analysis, rarely invoking compilation or simulation, which keeps their patches minimal and focused but leaves them vulnerable to semantic edge cases that would be caught by a build-test loop. Opus and Sonnet share the highest File-Level Precision in \autoref{tab:main_results}, yet their most common failure mode is under-validation of cross-module handshakes and stale fields that only manifest at runtime. GPT-5.4 (Codex CLI) is the most systematic explorer, performing heavy \texttt{rg}/shell sweeps and moderate build probing to gather broad context before editing. Its failures are typically not from lack of effort but from hidden coupling that survives apparently solid localization, and its patches tend to be larger than necessary as a result.

Among open-source models, GLM 5.1 achieves surprisingly strong results (63.1\% overall) through a symbol-driven, narrow reading style with light compile checks. However, its failed patches tend to target the correct subsystem but operate at the wrong abstraction layer. Qwen3.6 Plus takes a grep-first, local approach; most of its failed patches touch only one file and represent incomplete RTL surgery rather than total mislocalization. However, when Qwen loses confidence on complex cases, it exhibits a noisy tail, producing over-broad patches on OpenTitan and Caliptra. DeepSeek V3.2 runs the most terminal-intensive workflow with frequent compile checks, which helps it solve cases requiring iterative refinement, but its patches are the noisiest among all models, often containing broad mechanical edits that still miss the exact contract edge. Kimi K2.5 tends to stay too context-local, repeatedly revisiting the same file and editing the same layer without stepping back to consider cross-module dependencies, making it particularly weak on integration and configuration bugs.

\paragraph{Repository-Specific Challenges.}
Beyond the project-scope and bug-type factors discussed in Section~\ref{sec:cross_repo}, each repository presents distinctive failure characteristics. On XiangShan, failures cluster at pipeline stage boundaries involving dispatch/issue metadata, LSU replay timing, and predictor redirects, where the visible symptom is often one cycle downstream from the root cause. On Ibex, the remaining hard cases are architectural rather than local: trap handling, CSR semantics, and LSU behavior require coordinated changes across the controller, core, and register file. CVA6 has relatively few failures, but they concentrate in subtle cache-control and MMU semantics where agents find the right file but apply the wrong constant or cache-state condition. Caliptra combines RTL bugs with integration scaffolding: UVMF templates, regression YAML, and generated collateral frequently need updating alongside the RTL fix.

\subsection{Agent Behavior Analysis}

\begin{table}[t]
\centering
\caption{Agent behavioral statistics (mean per task). Tokens in thousands (K). Cost estimated from official API pricing.}
\label{tab:agent_behavior}
\resizebox{\columnwidth}{!}{%
\begin{tabular}{@{}l rrr r r@{}}
\toprule
\textbf{Model} & \textbf{Prompt (K)} & \textbf{Compl. (K)} & \textbf{Cache (\%)} & \textbf{Tool Calls} & \textbf{Cost (\$)} \\
\midrule
\multicolumn{6}{l}{\textit{Proprietary}} \\
GPT-5.4 xhigh    & 4,201 & 22.3 & 95.5 & 77.7 & 1.81 \\
Opus 4.6 high     & 756   & 4.9  & 99.7 & 20.0 & 0.51 \\
Sonnet 4.6 high   & 714   & 2.3  & 99.7 & 18.6 & 0.26 \\
\midrule
\multicolumn{6}{l}{\textit{Open-Source}} \\
GLM 5.1           & 1,605 & 15.9 & 97.1 & 40.3 & 0.54 \\
Qwen3.6 Plus      & 1,686 & 21.6 & ---  & 41.1 & 0.50 \\
DeepSeek V3.2     & 2,560 & 24.1 & 97.2 & 56.5 & 0.10 \\
Kimi K2.5         & 1,128 & 8.0  & 95.6 & 27.7 & 0.16 \\
\bottomrule
\end{tabular}%
}
\end{table}

\autoref{tab:agent_behavior} quantifies the behavioral differences observed in the failure analysis. GPT-5.4 xhigh averages 4.2M input tokens and 77.7 tool calls per task, consistent with its systematic shell-intensive exploration. The Claude models operate with much lower overhead: Sonnet averages 714K input tokens and 18.6 tool calls, achieving comparable resolved rates through the static analysis approach described above. DeepSeek V3.2 consumes 2.6M input tokens with 56.5 tool calls, reflecting its terminal-intensive debugging style.

Cost estimates based on official API pricing range from \$0.10 per task (DeepSeek V3.2) to \$1.81 (GPT-5.4 xhigh). Although open-source models offer substantially lower per-token pricing, their higher token consumption offsets much of this advantage: GLM 5.1 and Qwen3.6 Plus both cost around \$0.50 per task, comparable to Claude Opus 4.6 (\$0.51) despite Opus's per-token price being 4--10$\times$ higher. The Claude models benefit from both lower token usage and near-perfect cache hit rates (99.7\%), the latter likely aided by Claude Code's optimized context management and prompt caching strategy. Qwen3.6 Plus shows no cache utilization (marked --- in \autoref{tab:agent_behavior}), likely due to its API endpoint configuration.

%% file: src/05_conclusion.tex

\section{Conclusion}
We presented \bench, the first large-scale, repository-level benchmark for evaluating LLM agents on real-world hardware bug repair tasks, comprising \numcases instances derived from real historical bug-fix pull requests across six repositories spanning Verilog/SystemVerilog and Chisel, with fully containerized, execution-grounded evaluation and a largely automated construction pipeline that enables efficient expansion to new repositories. Our evaluation of seven LLMs with four agent frameworks shows that the best agent resolves 70.7\% of tasks overall, with performance exceeding 90\% on smaller cores but dropping below 65\% on complex SoC-level projects, a gap better explained by project scope and bug-type distribution than by code size alone. We also observe larger performance gaps across models than commonly reported on software benchmarks. Our failure analysis identifies three recurring failure modes, fault localization, hardware-semantic reasoning, and cross-artifact coordination, as well as distinctive model-specific behavioral differences and repository-specific challenges. We believe \bench can serve as a foundation for measuring and driving progress in LLM-assisted hardware engineering.